\documentclass{article}
    \usepackage{arxiv}
    \usepackage{moreverb,url}
    \usepackage[hidelinks]{hyperref}
    \usepackage{IEEEtrantools}
    \usepackage[square,numbers]{natbib}
    
    \def\imwidth{12cm}
    \def\lacol{7cm} % large column in table 1

\usepackage[utf8]{inputenc}
\usepackage[pdftex]{graphicx}
\usepackage{amsmath,stmaryrd,amssymb}
\usepackage{booktabs}
\usepackage{multirow}
\usepackage[table,xcdraw]{xcolor}
\usepackage{diagbox}

\def\eg{e.g.,\ }
\def\Eg{E.g.,\ }
\def\cf{\textit{cf.}\ }
\def\ie{\textit{i.e.,\ }}

\DeclareMathOperator{\Card}{Card}

\newcommand{\titleName}{
Fully Learnable Deep Wavelet Transform for Unsupervised Monitoring of High-Frequency Time Series}

\newcommand{\abstr}{%250 words
High-Frequency (HF) signals are ubiquitous in the industrial world and are of great use for monitoring of industrial assets. Most deep learning tools are designed for inputs of fixed and/or very limited size and many successful applications of deep learning to the industrial context use as inputs extracted features, which is a manually and often arduously obtained compact representation of the original signal. 
In this paper, we propose a fully unsupervised deep learning framework that is able to extract a meaningful and sparse representation of raw HF signals. 
We embed in our architecture important properties of the fast discrete wavelet transformation (FDWT) such as (1) the cascade algorithm, (2) the conjugate quadrature filter property that links together the wavelet, the scaling and transposed filter functions, and (3) the coefficient denoising. Using deep learning, we make this architecture fully learnable: both the wavelet bases and the wavelet coefficient denoising are learnable. To achieve this objective, we propose a new activation function that performs a learnable hard-thresholding of the wavelet coefficients. 
With our framework, the denoising FDWT becomes a fully learnable unsupervised tool that does neither require any type of pre- nor post-processing, nor any prior knowledge on wavelet transform.
We demonstrate the benefits of embedding all these properties on three machine-learning tasks performed on open source sound datasets. We perform an ablation study of the impact of each property on the performance of the architecture, achieve results well above baseline and outperform other state-of-the-art methods.
}

\newcommand{\keywo}{Fast Discrete Wavelet Decomposition; Deep Learning; High-Frequency signals; Unsupervised Anomaly Detection; Sparse Decomposition; Denoising }

\newcommand{\aknow}{
This work was supported by the Swiss National Science Foundation (SNSF) Grant no. PP00P2-176878 and by the Innosuisse grant no. 27662.1 PFES-ES.
The authors would like to thank Christoph Preisinger for his preliminary explorations of the proposed methodology.
}

\begin{document}

    \title{\titleName}
    \author{%
    \ifdefined\DoubleBlind
    \ 
    \else
        Gabriel Michau\\
        ETH Z\"urich,\\
        Z\"urich, Switzerland\\
        \And 
        Gaetan Frusque\\
        ETH Z\"urich,\\
        Z\"urich, Switzerland\\
        \And 
        Olga Fink\\
        ETH Z\"urich,\\
        Z\"urich, Switzerland
    \fi}
    \date{\today}
    \subtitle{Preprint}
    \rhead{
    \ifdefined\DoubleBlind
        \scshape Fully Learnable Deep Wavelet Transform
    \else
        \scshape G. Michau et al. - Fully Learnable Deep Wavelet Transform
    \fi}
    \ifdefined\NMI
        \date{}
        \subtitle{}
    \fi
        \maketitle
        \begin{abstract}
        \abstr
        \end{abstract}

%%%%%%%%%%%%%%%%%%%%%%%%%%%%%%%%%%%%%%%%%%%%%%%%%%%%%%%%%%%%%%%%%%%%%%%%%
\section{Introduction}
\label{sec:intro}
%%%%%%%%%%%%%%%%%%%%%%%%%%%%%%%%%%%%%%%%%%%%%%%%%%%%%%%%%%%%%%%%%%%%%%%%%

M%
onitoring of industrial assets often relies on high-frequency (HF) signal measurements, such as electric current, vibrations or sound.
One difficulty of dealing with such signals in the industrial context is the conciliation between the high-frequency sampling and low dimensional decision states (\eg healthy/unhealthy), in a context where, very often, labels are not available. Therefore, many industrial applications require unsupervised approaches able to extract meaningful and sparse information from HF signals, in order to ease the process analysis, the diagnostics and more generally the optimisation of the assets' life cycles. 

Before the recent developments of large storage capacity and high computational powers, raw HF signals could not be recorded, enforcing companies to spend time and budget on devising relevant features for later analysis, achieving in that way a sparse representation of the input data. These features could be of various nature, such as spectral features, based on the Fourier Transform, the Fast Fourier Transform, or on wavelets~\cite{peng2004application}, on statistical features (moments, energy, entropy, etc.), or on descriptive features (envelopes, amplitude, etc.)~\cite{gangsar2020signal}.

In recent days, storing HF data has become less of a technical problem, and handling large datasets efficiently has been made possible with the rise of deep learning. However, most deep learning tools are designed for inputs of fixed and/or very limited size. Many successful applications of deep learning to industrial context use as inputs extracted features, that is, a manually obtained compact representation of the original signals. Very often these features are a spectrogram~\cite{wang2020missing}, wavelet coefficients statistics~\cite{li2019deep}, or others~\cite{fawaz2019deep,wang2019time,fink2020potential}. Although such frameworks are extremely successful, they still require careful feature extraction with the right hyper-parameters, which can be a time consuming task. 
In addition, the extracted features might be sensitive to unexpected noise or to changing conditions, and the design of domain invariant unsupervised features, whether with post-processing or with deep learning, is still an open research question~\cite{michau2021unsupervised}.  

With the development of convolution neural networks~\cite{kiranyaz20211d}, early works realised that temporal CNN is equivalent to a digital filter and that it could learn convolution kernels similar to a Fourier transform or to wavelets~\cite{zhang2018deep} or also be used to learn sparse representations~\cite{papyan2017convolutional}. It was soon proposed to constrain the network to perform operations similar to the Fourier Transform~\cite{liu2013fourier,uteuliyeva2020fourier}, or to wavelet transform either with continuous wavelet transform ~\cite{luan2018gabor,li2021waveletkernelnet} or with discrete wavelet transform~\cite{wang2018multilevel,recoskie2018learning,liu2019multi,khalil2020end}. 
All these works demonstrate that by using architectures or kernels inspired by spectral analysis, superior results could be obtained on supervised deep learning tasks. Yet, these approaches are rarely adapted to unsupervised machine learning tasks and the link with the spectral transformation is often restrained either to the network architecture only, or to the initialisation of the convolution kernels. In addition, Fourier-Transform-based deep learning architectures become rapidly too heavy to handle when the size of the input time series increases.  

To mitigate the above mentioned limitations, we propose in this work a new deep learning framework based on the Fast Discrete Wavelet Transformn (FDWT) that allows an automatic and easy extraction of meaningful and sparse representation of the input signals. First, we propose here to mimic the FDWT cascade architecture utilising the deep learning framework. We, thus, propose to learn at each decomposition level, on the one hand the right high- and low-pass filters, and on the other hand the right hard-thresholding coefficients for denoising. Second, for the learning of the filters, we take advantage of the long-known advantageous properties of orthogonal wavelet filter banks with the Conjugate Quadrature Filter property to structure the learning process while making sure the final network remains similar to a FDWT operation. It also makes very light architecture with only few parameters to learn (few hundreds). This is opposite to the general trend in deep learning to learn millions or billions of parameters. Third, to learn the right hard-thresholding operation, we propose a new learnable activation function which is continuous and differentiable and which approximates the hard-thresholding operation. It can thus be used as an integral part of the deep learning architecture and it removes the need for human analysis and decisions on these difficult tasks~\cite{donoho1994threshold}.

After presenting in Section~\ref{sec:FDWT} important properties and characteristics of the FDWT, we show how to translate these properties into a deep neural network in Section~\ref{sec:methods}.  In Section~\ref{sec:experiments}, we test our approach on three tasks, one classification task and two unsupervised anomaly detection tasks. Starting from the FDWT, we demonstrate how each of our different contribution contributes toward better results, well above baseline.
Last, in Section~\ref{sec:compOther} we compare our approach to several other architectures, such as the Scattering Transform~\cite{anden2014deep}, U-net~\cite{jimenez2019u} and a convolution auto-encoder. We show that without the need of any pre-processing step, with our particularly light architecture, we achieve very competitive results.

%%%%%%%%%%%%%%%%%%%%%%%%%%%%%%%%%%%%%%%%%%%%%%%%%%%%%%%%%%%%%%%%%%%%%%%%%
\section{Background on the Cascade and FDWT}
\label{sec:FDWT}
%%%%%%%%%%%%%%%%%%%%%%%%%%%%%%%%%%%%%%%%%%%%%%%%%%%%%%%%%%%%%%%%%%%%%%%%%

\begin{figure*}
    \centering
    \includegraphics[width=0.7\textwidth]{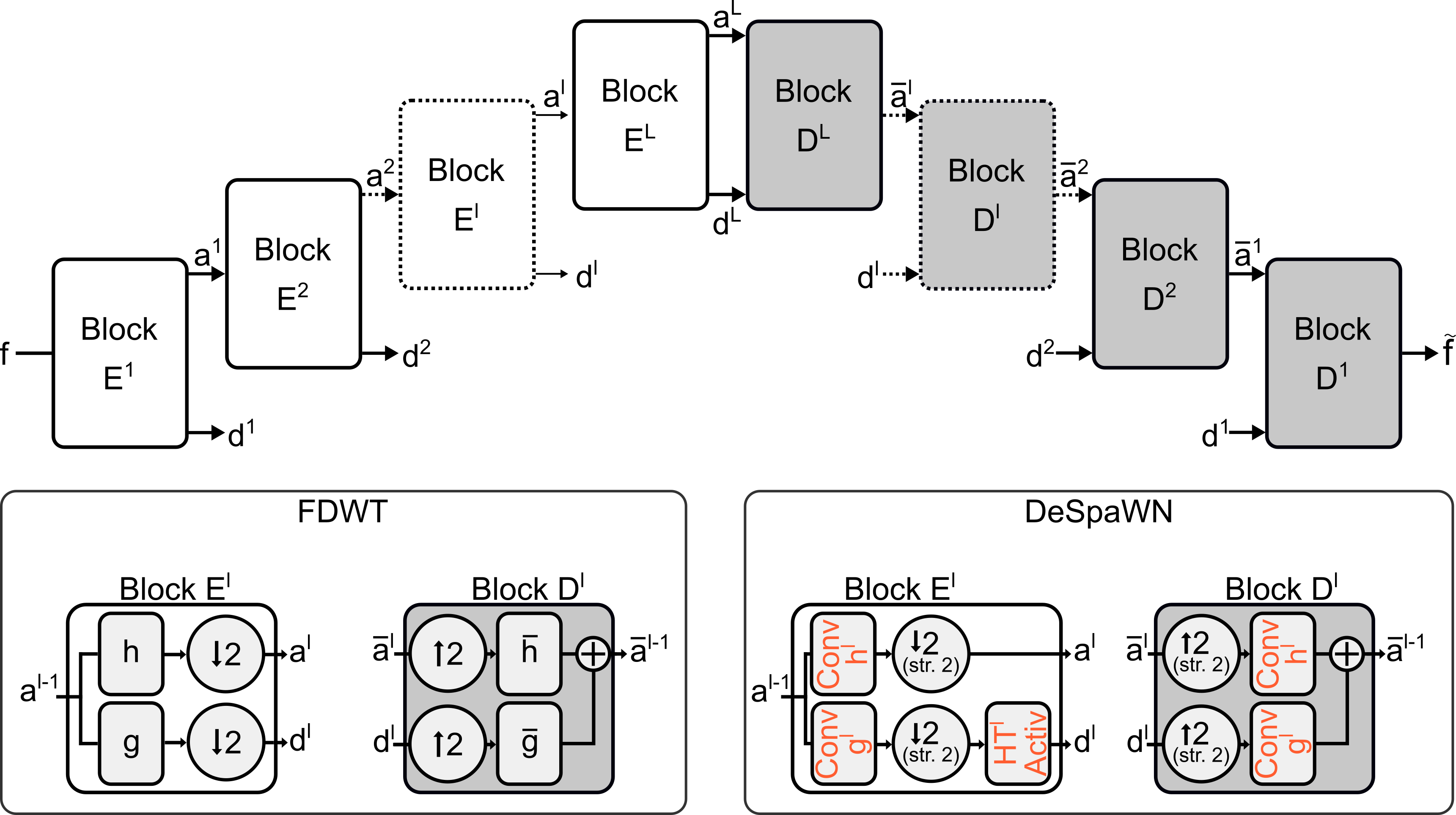}
    \caption{\textbf{Cascade - FDWT versus DeSpaWN}. The FDWT can be modelled as a convolution neural network, an auto-encoder of $2L$ layers, comprising $L$ encoding blocks and $L$ decoding blocks. Each encoding block has two outputs, one of which is connected to the corresponding decoding block with skip-connections. In this work, we propose to make the network learnable, that is, to learn the right filters. In addition, we propose a new learnable hard-thresholding activation function which allows to learn the wavelet coefficient denoising operation at the same time. Red elements in the Figure are learnable. The proposed architecture mimics the Denoising FDWT and is denoted as Denoising Sparse Wavelet Network (DeSpaWN).
    \label{fig:fullArchitecture}}
\end{figure*}

    \subsection{Cascade algorithm}
    %----------------------------------------------------------------------

The fast discrete wavelet transform (FDWT) uses wavelets designed such that the family of wavelets made out of their scaling and translation by any power of $2$ makes an orthonormal family~\cite{mallat2009wavelet}. Using such wavelets, and their corresponding scaling function, FDWT decomposes successively each approximation of a signal $f$ into a coarser approximation $a$ (low-pass filtered version of the signal $f$) and its detail coefficients $d$ (also denoted as wavelet coefficients, high-pass filtered version of $f$). With an orthonormal basis of $\mathbf{L}^2(\mathbb{R})$, the decomposition of any function with such a transformation is invertible. Additionally, one can demonstrate that, due to the factor 2 in scale between the levels and in translation between the coefficients, the decomposition at level $l$ can be expressed as a function of the previous approximation, sub-sampled by a factor $2$ and the original non-dilated wavelet (interested readers are referred to \cite[Chapter 7.3.1]{mallat2009wavelet} for the proof). Similarly, for the reconstruction at each level, it can be expressed as the convolution of the conjugate of the original wavelet with the previous level reconstruction, where zeros have been interpolated between every sample. FDWT uses this property to apply, in cascade, the exact same operation at each level, using a single wavelet, but down-scaling and interpolating with zeros the signals at each level of the decomposition and the reconstruction respectively. It is denoted as \emph{the cascade algorithm}. Figure~\ref{fig:fullArchitecture} illustrates the FDWT cascade algorithm, where $g$ is the wavelet, $h$ is the scaling function, $\bar{g}$ and $\bar{h}$ their respective conjugate filters.

    \subsection{CQF properties}
    %----------------------------------------------------------------------
An interesting property of the fast discrete wavelet transform has been discovered in 1976 by Croisier, Esteban, and Galand~\cite{croisier1976perfect,mallat2009wavelet} and was extended in 1984 in~\cite{smith1984procedure,mintzer1985filters,mallat2009wavelet}. It establishes the ground for finding filters allowing a perfect reconstruction of the input signal as using the \emph{Conjugate Quadrature Filter} (CQF) bank property. The quadrature property ensures a symmetric response of the decomposition filters with respect to the cut-off frequency and ensures thus an anti-aliasing property. To do so, the filters can be designed such that the wavelet function $g$ is the alternative flip of the scaling function $h$. The conjugate property ensures that the reconstruction filters have an anti-cancellation property. Both properties combined are usually denoted as CQF which is formalised as follows:
\begin{equation}
\label{eq:CQF}
\left\lbrace
\begin{array}{rl}
        g[n] & = (-1)^n\cdot h[-n],\\
        \bar{h}[n] & = h[-n],\\
        \bar{g}[n] & = (-1)^{(n+1)}\cdot h[n],\\
    \end{array}
\right.
\end{equation}
where $h[-n]$ denotes the $n$th coefficient of $h$ in reversed order.

To achieve a perfect reconstruction, the frequency content conservation after applying the filters imposes that the sum of the responses of both filters should be $2$, which imposes further constraints on the filters orthonormality~\cite{mallat2009wavelet}, usually considered as part of the CQF properties.

    \subsection{Signal denoising with FDWT}
    %----------------------------------------------------------------------
One of the major applications of the wavelet analysis is signal denoising (\eg \cite{alfaouri2008ecg,bayer2019iterative}). The usual assumption is that regular and structured signals, once decomposed under the right wavelet basis, will naturally lead to a sparse decomposition~\cite{mallat2009wavelet}. They will activate only certain wavelet coefficients at specific time and decomposition levels. As a consequence, noise, by nature unstructured, activates the wavelet filters at any level, but usually with a small amplitude. Thus, denoising usually consists in applying a hard thresholding function to the resulting wavelet coefficients before applying the reconstruction algorithm~\cite{donoho1993nonlinear}. 
However, finding the right thresholding parameters is a difficult task that has been the topic of extensive research~\cite{donoho1994threshold}.

%%%%%%%%%%%%%%%%%%%%%%%%%%%%%%%%%%%%%%%%%%%%%%%%%%%%%%%%%%%%%%%%%%%%%%%%%
\section{Learnable Denoising Sparse Wavelet Network}
\label{sec:methods}
%%%%%%%%%%%%%%%%%%%%%%%%%%%%%%%%%%%%%%%%%%%%%%%%%%%%%%%%%%%%%%%%%%%%%%%%%

    \subsection{Architecture overview}
    %----------------------------------------------------------------------

In this research, we propose the Denoising Sparse Wavelet Network (DeSpaWN). DeSpaWN is utilising a fully learnable cascade network, mimicking an $L$-level wavelet cascade, such as illustrated in Figure~\ref{fig:fullArchitecture}. It, thus, consists of $L$ encoding blocks, and $L$ decoding blocks. Each encoding block $l$ is composed of two learnable convolutional layers $g^l$ and $h^l$ with a stride of two, analogous to the wavelet and scaling filters with down-sampling in the FDWT. The resulting coefficients are fed to a specifically designed learnable hard thresholding layer $HT^l$, which is similar to wavelet denoising operation, and to the next block for further decomposition. Similarly to the FDWT, each decoding block takes two inputs, the coefficients from the previous block and the detail coefficients from its corresponding encoding block $l$. It applies to each input a learnable convolution transpose layer $\bar{g}^l$ and $\bar{h}^l$ with a stride of two and sums the results of both layers together. We designed the deep neural network with two main distinctive properties: First, all convolution kernels and second, all positive and negative hard-thresholds, are learnable. This makes possible to learn fully and in a completely unsupervised way, the most adapted denoising FDWT for the input signals.

In addition, according to the work in~\cite{bolcskei2019optimal}, a sparsely connected deep neural network, such as the one proposed here, is able to approximate representation systems that encompasses and are more general than the representation system provided by the wavelet.

Overall, with the proposed architecture, the network has $(k_n+2)\cdot L$ learnable parameters, where $k_n$ is the number of coefficients of the wavelets. For example, mimicking Daubechies-4 wavelets, $k_n$ is set to $8$ and the network has thus $10\cdot L$ learnable parameters. Since $L$ cannot be set larger than the nearest second logarithm of the training input size, the number of parameters is unlikely to exceed few hundreds.

    \subsection{Objective function}
    %----------------------------------------------------------------------
In this work, we propose to learn the best wavelet and scaling functions for achieving a sparse decomposition. This means that we have two learning objectives: first, a good signal reconstruction and second, a sparse decomposition. Sparsity is usually measured by achieving the smallest $\ell_0$-norm of the resulting coefficients. Yet, this is a non-convex metric. A typical convex surrogate of the $\ell_0$-norm is the $\ell_1$-norm. Part of our objective function should be designed such as to minimise the $\ell_1$-norm of the obtained wavelet coefficients. As a consequence, we propose that for the second part of our objective function, we also measure the $\ell_1$-norm of the reconstruction error such as to make the two terms comparable.

We, thus, train our network with the following objective function:
\begin{equation}
\mathcal{L} = \frac{1}{\Card(f)}\vert f - \tilde{f} \vert_1 + \gamma \cdot \mathbf{L}\left(\lbrace\mathbf{d^l}\rbrace_{l\in [1..L]},\mathbf{a^L}\right)
\label{eqn:loss}
\end{equation}
where the first part of the loss is the averaged $\ell_1$-norm of the residuals on the reconstruction and the second part is the sparsity term, here proposed as the average of all wavelet coefficient modulus and of the last layer approximation coefficient modulus. 
\begin{equation}
\resizebox{\ifdim\width>\columnwidth
            \columnwidth
          \else
            \width
          \fi}{!}{$
\mathbf{L}\left(\lbrace\mathbf{d^l}\rbrace_{l\in [1..L]},\mathbf{a^L}\right) = \frac{1}{\Card(\lbrace\mathbf{d^l}\rbrace_{l\in [1..L]},\mathbf{a^L})} \sum_{l\in[1..L]} \vert \mathbf{d^l}\vert_1 + \vert \mathbf{a^L} \vert_1$.}
\end{equation}

    \subsection{CQF-constrained architecture}
    %----------------------------------------------------------------------
As seen in the previous section, the CQF property for the wavelet filters ensures perfect and anti-aliasing reconstruction. We, thus, propose to utilise this property, at the same time to simplify the learning process and to harness its advantages. As a matter of fact, using the CQF property as defined in \eqref{eq:CQF}, learning a single kernel would define the other three. The second advantage is, that, based on the above idea of penalising the $\ell_1$ norm of the wavelet coefficient as a surrogate for the non-convex $\ell_0$-norm, the constraint-free learning of the kernels might lead toward a state where the coefficients of $g$ and $h$ are minimised to result in a small $\ell_1$ value in the latent space, while the reconstruction is still possible thanks to large coefficients in $\bar{g}$ and $\bar{h}$, which goes against the original goal of achieving sparsity. Constraining the values of $\bar{g}$ and $\bar{h}$ based on those of $g$ and/or $h$ mitigates this issue.

One option would be to learn a single kernel $h^0$, use the CQF constraints to derive $g^0$, $\bar{g}^0$ and $\bar{h}^0$ and then, impose that all layers of the network use the same kernels, mimicking in that way the traditional wavelet decomposition with a single wavelet basis. However, we state that this approach would not benefit from the full potential of deep learning. 

Alternatively, we propose to learn one kernel per layer $\left\lbrace h^l\right\rbrace_{l\in [1..L]}$ and to use the CQF property to constraint the other three kernels of the layer ($g^l$, $\bar{h}^l$, $\bar{g}^l$). A similar alternative would be to learn independently the low-pass $h^l$ and high-pass $g^l$ filters but impose the partial CQF constraints such that $\bar{h}^l[n] = h^l[-n]$ and $\bar{g}^l[n] = g^l[-n]$. These two types of architectures learn at each level the best wavelets to describe the signal at this scale in a sparse way.

We implement all these different variations of the CQF property and compare them in Section~\ref{sec:experiments}.

Last, with the proposed architecture we already cover our three training objectives: a good and sparse reconstruction thanks to the objective function, and a stable learning thanks to the constraints imposed between decomposition and reconstruction filters. Therefore, we do not try to impose further constrains on the filters, in particular not the orthonormality property.

    \subsection{Learnable Denoising}
    %----------------------------------------------------------------------

One assumption behind the wavelet decomposition is that a structured signal should lead to sparse coefficients under the right wavelet basis. This is the property we encourage by minimising the $\ell_1$-norm of the wavelet coefficients in our objective function. However, the addition of noise to the input signal, which is by nature non-structured, would lead to a random activation of the filters, independently of the chosen filters. This would, therefore, necessarily lead to a non-sparse decomposition. The sparsity of the decomposition is, thus, sought only once the noise has been canceled. This problem is usually tackled under the assumption that noise would lead to small activation of the filters. As a consequence, the impact of noise on the decomposition can be removed by hard-thresholding the obtained coefficients. In \cite{donoho1994ideal}, guarantees are provided for recovering and denoising signals observed in Gaussian noise by applying the right hard-thresholding operation. Yet, finding the right thresholding parameters is a difficult task and usually depends on the use-case and the specific dataset. 

\begin{figure}
    \centering
    \includegraphics[height=6cm]{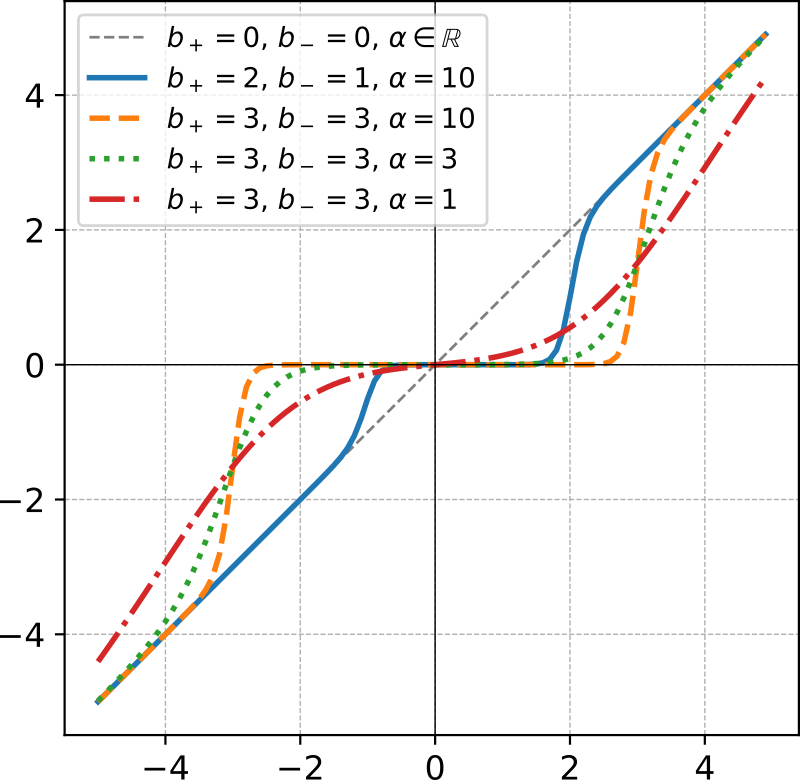}
    \caption{\textbf{Hard Thresholding Activation Function}. Proposed Activation Function performing an operation close to an asymmetric hard thresholding. Both thresholds, in the negative and positive half-space are learnable parameters. In our paper, $\alpha$ is set to $10$ to simulate a ``sharp'' thresholding. 
    \label{fig:HT}}
\end{figure}

In this paper, we propose to make the thresholding step part of our architecture to learn the best thresholding parameters and to remove the need of handling it as a separate step. We introduce a learnable hard-thresholding activation function as a combination of two opposite sigmoid functions:
\begin{equation}
\resizebox{\ifdim\width>\columnwidth
                \columnwidth
              \else
                \width
              \fi}{!}{$
    \forall x \in \mathbb{R},\quad HT(x) = x\left[\frac{1}{1+\exp\left(\alpha\cdot\left(x+b_-\right)\right)}+\frac{1}{1+\exp\left(-\alpha\cdot\left(x-b_+\right)\right)} \right]
    \label{eqn:fullHT}$}
\end{equation}
where $\alpha$ is a ``sharpeness'' factor arbitrarily fixed to $10$ in this paper, $b_+$ and $b_-$ are the positive and negative learnable bias acting as the thresholds on both sides of the origin, as illustrated in Figure~\ref{fig:HT}.

Denoting the sigmoid function $1/(1+\exp(-x))$ by $\sigma$, \eqref{eqn:fullHT} becomes 
\begin{equation}
    \forall x \in \mathbb{R},\quad HT(x) = x\cdot \left[ \sigma\left(-\alpha\cdot\left(x+b_-\right)\right) + \sigma\left(\alpha\cdot\left(x-b_+\right)\right) \right]
\end{equation}

To replicate the original FDWT without denoising, one can fix in this layer, $b_+$ and $b_-$ to zero, enforcing, thereby, a linear activation.

%%%%%%%%%%%%%%%%%%%%%%%%%%%%%%%%%%%%%%%%%%%%%%%%%%%%%%%%%%%%%%%%%%%%%%%%%
\section{Comparison between traditional FDWT and DeSpaWN}
\label{sec:experiments}
%%%%%%%%%%%%%%%%%%%%%%%%%%%%%%%%%%%%%%%%%%%%%%%%%%%%%%%%%%%%%%%%%%%%%%%%%

    \subsection{Machine Learning for Sound Data}
    %----------------------------------------------
We focus our experiments specifically on sound data as it is a fast growing field in industrial applications~\cite{liebetrau2017predictive,li2019deep}. Sound-based analysis raises the interest of more and more companies for several reasons: experts and machine operators already listen to the machines and are able to tell when the operation is not nominal. It should, therefore, be possible to train a machine learning approach for continuous monitoring. Moreover, industrial processes are inherently noisy. Thus, it is expected that their sound contains information on their process state. Since the sound is by nature multi-scale, it might also allow for the monitoring of several systems at once. Monitoring industrial machines with sound is also relatively simple and cheap. Hence, it is an attractive and scalable solution. Microphones are easy to install or retrofit. They are not intrusive. Hardware and software are readily available. Yet, sound data come with their own challenges such as high noise levels due to the machines usually operated in factory environments, hence, in noisy environments. Besides, finding how to relate specific industrial processes to the recorded sound is a difficult task. Automatic and noise-robust handling of sound data is, therefore, of high interest for many industrial applications.

    \subsection{Classification and Anomaly Detection}
    \label{sec:classifAnomAppro}
    %-----------------------------------------------------------

The objective of DeSpaWN is  learning of a robust auto-encoder. This robustness is achieved through denoising with the hard-thresholding of the learned coefficients, and through sparsity, forcing the network to learn the most meaningful wavelet to describe the training data. Thus, we state that once trained, the auto-encoder should produce for similar signals, similarly sparse latent space, and achieve similarly low reconstruction residuals. We use these properties to design our classification and our anomaly detection strategies.

For classification, we propose an approach similar to dictionary learning, which consists in training one auto-encoder per class and to assign to a sample the class corresponding to the auto-encoder that led to minimal loss of the objective function in \eqref{eqn:loss}. This classification approach is common in signal processing when signal models are learned or trained to capture very specific properties of a class~\cite{mairal2008discriminative,recoskie2018learning}. 

For anomaly detection, we state that anomalies can be broadly separated into two types: local intermittent anomalies in the signal (an abnormal pulse due to an impact, \eg a broken tooth in a gear box) or trend anomalies, when the signal behaviour changes more globally (\eg a change of frequency due to increased friction). When dealing with long signals (several seconds to minutes), capturing trend anomalies is usually achieved by looking at global indicators, such as the network objective function. However, such approaches tend to hide local anomalies, which are averaged out when the signal length increases. One possible solution can be shortening of the input time series to mitigate the impact of the average. However, this solution is not always practical. Here, we propose instead to jointly use at averaged and local statistics. More precisely, per time series we propose to use the average residual ($Res$), the maximum residual($MaxRes$), the average sparsity loss (average $\ell_1$-norm of the coefficients per level, $\lbrace\hat{\ell}^l_1\rbrace_{l\in[1..L]}$ ), and the maximum sparsity loss ($\lbrace\bar{\ell}^l_1\rbrace_{l\in[1..L]}$). We then apply a one-class classifier on this new latent space, $Res-MaxRes-\lbrace\hat{\ell}^l_1\rbrace-\lbrace\bar{\ell}^l_1\rbrace_{l\in[1..L]}$, of size ($2+2\cdot L$). 

In this work, we achieved similar results using either of the one-class Isolation Forest, the once-class Support Vector Machine, an Elliptic Envelope or a one-class Extreme Learning Machine (ELM)~\cite{michau2020feature}. We, therefore, report only results using the later, an ELM with 50 neurons.

    \subsection{Datasets}
    %----------------------------------------------
We test the proposed approach on two open source sound datasets. First, we test the model on an anomaly detection task of sound data of industrial machines with the Sound Dataset for Malfunctioning Industrial Machine Investigation and Inspection (MIMII)~\cite{purohit2019mimii}. Second, we demonstrate that the network can learn decomposition specific to its training data by solving a dictionary learning classification task. We show that the approach performs equally well independently of the type of sound used as input to perform classification on the bird song dataset, as proposed in~\cite{recoskie2018learning}. Last, we show the consistency of the results by using this same dataset for anomaly detection. For both datasets, we analyse the obtained latent space and demonstrate that it can also be used to interpret the data at hand.

        \subsubsection{Malfunctioning Industrial Machine Detection}
        %```````````````````````````````````````````````````````````
\begin{table}
\centering
\caption{\textbf{Characteristics of the MIMII Dataset}\label{tab:mimii}}
\begin{tabular}{@{}lrrrp{\lacol}@{}} % p{4cm}
\toprule
Type & ID & Normal & Abn. & Operating Conditions \& Type of Anomalies                                           \\ \midrule
Valve        & 0          & $991$  & $119$    & \multirow{4}{=}{\parbox{\lacol}{Open / close repeat with different timing.\\More than two kinds of contamination.}} \\
             & 2          & $708$  & $120$                                                                &                                                             \\
             & 4          & $1000$ & $120$                                                                &                                                             \\
             & 6          & $992$  & $120$                                                                &                                                             \\\midrule
Pump         & 0          & $1006$ & $143$    & \multirow{4}{=}{\parbox{\lacol}{Suction from / discharge to a water pool.\\Leakage, contamination, clogging, etc.}}\\
             & 2          & $1005$ & $111$                                                                &                                                            \\
             & 4          & $702$  & $100$                                                                &                                                             \\
             & 6          & $1036$ & $102$                                                                &                                                             \\\midrule
Fan          & 0          & $1011$ & $407$    & \multirow{4}{=}{\parbox{\lacol}{Normal operation.\\Unbalanced, voltage change, clogging, etc.}}\\
             & 2          & $1016$ & $359$                                                                &                                                             \\
             & 4          & $1033$ & $348$                                                                &                                                             \\
             & 6          & $1015$ & $361$                                                                &                                                             \\\midrule
Slide Rail   & 0          & $1068$ & $356$    & \multirow{4}{=}{\parbox{\lacol}{Slide repeat at different speeds.\\Rail damage, loose belt, no grease, etc.}}\\
             & 2          & $1068$ & $267$                                                                &                                                             \\
             & 4          & $534$  & $178$                                                                &                                                             \\
             & 6          & $534$  & $89$                                                                 &                                                             \\ \bottomrule
\end{tabular}
\end{table}

The MIMII dataset, also known as Sound Dataset for Malfunctioning Industrial Machine Investigation and Inspection~\cite{purohit2019mimii} consists of audio-recordings of four types of industrial machines, \ie valves, pumps, fans, and slide rails, in normal and malfunctioning states. It is, therefore, a good benchmark for testing anomaly detection approaches on sound data.

The dataset has four individual machines of four machine types. For each machine, sound from normal and abnormal operating conditions has been recorded without further label on the operating state or on the faults, making the dataset very suitable for unsupervised anomaly detection. Each machine has been recorded under three different Signal-to-Noise-Ratio (SNR) setups (0dB, 6dB and -6dB), where the noise denotes background noise of other industrial processes. This results in 48 experiments on which anomaly detection can be performed. It is interesting to note that various anomaly types are collected, and that several anomalies can influence the same machine in different recording samples. Table~\ref{tab:mimii} gives an overview of the dataset: presenting the number of samples for each machine and the conditions of operation.

The data have been recorded with a 8-channel microphone array, at 16kHz and 16bits resolution. Each sample is 10 seconds long, or 160 000 timestamps. With this, we can set $L$ to 17. In accordance with the work in~\cite{purohit2019mimii}, the data recorded by the first microphone only are used.

        \subsubsection{Bird Song Dataset}
        %```````````````````````````````````````````````````````````
For the second experiment, we use a different type of sound data to demonstrate the performance of the proposed framework in a different context, especially since industrial machines often make repetitive noise that is rather easier to characterise. The recordings of birdsongs in their natural environment are also subject to environmental noise and differences in the recording hardware that also influences the recorded sound. Since the data are labelled (contrary to the machine sound data), it allows us to test the proposed architecture both in an anomaly detection and in a classification setup.

The Xeno-canto Foundation collection bird song dataset~\cite{xeno2004dataset} is a dataset gathering bird songs from all around the world and collected by a large variety of participants. In~\cite{recoskie2018learning}, the author proposes to focus on the following birds: corn bunting (CB), Eurasian  skylark (ES),  barn  swallow (BS),  sedge  warbler (SW),  and  common  nightingale (CN). These species were selected under the argument that all their recordings were recorded by the same person, implying similar recording conditions and probably similar and consistent hardware. This allows crossing out the hypothesis that detected fluctuations between recordings and bird species would be due to a change of recording hardware.

For each of the above species, three recordings of about 5 minutes are available. To establish a fair comparison, we apply the exact same pre-processing as in \cite{recoskie2018learning}: decimation of the signals by a factor 4 since most of the signal energy is below 5kHz and the original sampling rate is 41kHz. The recordings are split into a collection of signals with $2^{18}$ samples ($\approx$24 second). This leads to the following number of signals:
\begin{itemize}
    \item Corn Bunting (CB):  30 signals
    \item Eurasian Skylark (ES):  24 signals
    \item Barn Swallow (BS):  21 signals
    \item Sedge Warbler (SW):  20 signals
    \item Common Nightingale (CN):  20 signals
\end{itemize}

As in~\cite{recoskie2018learning}, a 5-fold cross validation is used, meaning that 80\% of the data are used for training and 20\% for testing at each fold.

    \subsection{Ablation Study}
    %----------------------------------------------
        \subsubsection{From FDWT to DeSpaWN}
        %```````````````````````````````````````````````````````````
In this section, we aim at analysing how the different contributions impact the results compared to the traditional case where the coefficients from the FDWT would be used as inputs to subsequent machine learning tasks, such as classification or anomaly detection. Thus, in this first evaluation, to demonstrate how all the steps of the transition from the traditional FDWT to our proposed framework impact the results, we compare the results for the following architectures: 
\begin{itemize}
    \item \textbf{\textit{db4}}: Using the \textit{db4} wavelets
    \item \textbf{\textit{db4}+HT}: Using the \textit{db4} wavelets with learnable hard-thresholding of the coefficients.
    \item \textbf{\textit{CWN}} (CQF Wavelet Network): Learning a single kernel $h^0$, using CQF to fix the other three $g^0$, $\bar{h}^0$ and $\bar{g}^0$, and using these kernels for all levels.
    \item \textbf{\textit{LCWN}} (Layer-wise CQF): Learning one kernel $h^l$ per level in $L$, using CQF to fix the others
    \item \textbf{\textit{DeCWN}}(Denoising CQF): Learning a single kernel $h^0$, using CQF to fix the other for all levels and using the learnable hard-thresholding activation function.
    \item \textbf{\textit{DeSpaWN}}: Learning one kernel $h^l$ per level in $L$, using CQF to fix the others and using the learnable hard-thresholding activation function.
    \item \textbf{\textit{DeSpaWN-2}}: Learning two kernels, $h^l$ and $g^l$, per level in $L$, using CQF to fix the other and using the learnable hard-thresholding activation function.
    \item \textbf{\textit{FreeWN}}: Learning all kernels of all levels independently and using the learnable hard-thresholding activation function.
\end{itemize}

For all experiments, we set $L$, the number of decomposition levels to the nearest 2nd-logarithm of the length of the time series. We set the kernel size to 8, and compare the results with those achieved using Daubechies db4 wavelets (since they also have 8 coefficients) and with the baseline results on these datasets. In \eqref{eqn:loss}, we set $\gamma$, the weight on the sparsity term in the loss arbitrarily to one. The architecture has, therefore, $(8+2)\cdot L$ learnable parameters (eight kernel coefficients, two thresholds).

For reference, we also report results from the baseline models (MIMII~\cite{purohit2019mimii}, Bird Song~\cite{recoskie2018learning}).

        \subsubsection{Results and discussion}
        %```````````````````````````````````````````````````````````
\begin{table*}
\centering
\caption{\textbf{Comparative Study on three machine learning tasks:} For the different architecture variations (one per column), comparative results on the three considered tasks. For the anomaly detection tasks, on MIMII and on the Bird Song, we report the average AUC (\%). Last, for the Bird Song classification by dictionary learning, we report the Average Accuracy (\%).%
\label{tab:FullComp}}
\begin{tabular}{@{}crrrrrrrrrrr@{}}
\toprule
         & \multicolumn{3}{c}{DeSpaWN}                                                                        & \multicolumn{1}{c}{}    & \multicolumn{1}{c}{db4} & \multicolumn{1}{c}{}    & \multicolumn{1}{c}{LC} & \multicolumn{1}{c}{DeC} & \multicolumn{1}{c}{DeSpa} & \multicolumn{1}{c}{Free} &  \multicolumn{1}{c}{Base}\\ 
         & \multicolumn{1}{c}{$\gamma=1$} & \multicolumn{1}{c}{$\gamma=0.5$} & \multicolumn{1}{c}{$\gamma=5$} & \multicolumn{1}{c}{db4} & \multicolumn{1}{c}{+HT} & \multicolumn{1}{c}{CWN} & \multicolumn{1}{c}{WN} & \multicolumn{1}{c}{WN}  & \multicolumn{1}{c}{WN-2}  & \multicolumn{1}{c}{WN} &  \multicolumn{1}{c}{line}\\ \midrule
         & \multicolumn{11}{c}{Anomaly Detection on MIMII}                                                                                                                                                                                                                                                      \\\midrule
Valve    & 92.8                           & 92.7                             & 91.0                           & 92.7                    & 92.7                    & 92.7                    & 92.7                   & 92.9                    & \textbf{93.0}                      & \textbf{93.0}                & 61.3    \\
Pump     & \textbf{84.5}                           & 82.0                             & 72.6                           & 77.9                    & 78.3                    & 78.2                    & 78.1                   & 78.0                    & 84.2                      & 75.8                & 72.3    \\
Fan      & \textbf{86.2}                           & 84.8                             & 84.9                           & 84.6                    & 84.7                    & 84.9                    & 84.8                   & 85.3                    & 85.5                      & 83.8                &  79.0   \\
Slider   & \textbf{91.0}                           & 89.4                             & 78.7                           & 89.8                    & 90.0                    & 89.4                    & 89.6                   & 89.7                    & 90.1                      & 89.3                & 78.6    \\
6 dB     & \textbf{94.6}                           & 93.5                             & 84.4                           & 93.4                    & 93.4                    & 93.7                    & 93.5                   & 93.5                    & 94.3                      & 90.7                & 81.6    \\
0    dB  & \textbf{90.5}                           & 87.7                             & 81.9                           & 82.6                    & 88.1                    & 87.7                    & 87.8                   & 88.5                    & 90.2                      & 87.0                & 72.3    \\
-6    dB & \textbf{80.8}                           & 79.6                             & 76.5                           & 77.6                    & 77.9                    & 77.5                    & 77.5                   & 77.5                    & 80.0                      & 78.8                 & 64.4   \\ \midrule
\textbf{Avg.}     & \textbf{88.6}                           & {87.1}                             & {81.4}                           & {85.5}                    & {86.4}                   & {86.3}                    & {86.3}                   & {86.5}                    & {88.2}                      & {85.5}              & {72.8}      \\ \midrule\midrule
         & \multicolumn{11}{c}{Anomaly Detection (1 versus 4 bird species)}                                                                                                                                                                                                                           \\\midrule
{Avg.}      & \textbf{99.8}                           & \textbf{99.8}                             & {91.7}                           & {95.4}                    & {98.2}                    & {97.5}                   & {98.8}                   & {99.0}                    & {99.5}                      & \textbf{99.1}              &  N.A.     \\\midrule\midrule
\#C      & \multicolumn{11}{c}{Classification with dictionary learning (BirdSong)}                                                                                                                                                                                                                                             \\\midrule
2        & \textbf{99.2}                           & 98.2                             & 91.3                           & 50.0                    & 87.0                    & 73.3                    & 92.7                   & 93.4                    & 99.1                      & 89.0                & 97.2    \\
3        & \textbf{98.3}                           & 96.3                             & 88.7                           & 33.3                    & 77.3                    & 55.1                    & 86.0                   & 87.4                    & 97.5                      & 78.0                & 88.0    \\
4        & \textbf{97.5}                           & 94.5                             & 87.0                           & 25.0                    & 70.0                    & 43.5                    & 80.0                   & 81.9                    & 97.3                      & 67.0                & 74.7    \\
5        & \textbf{96.7}                           & 92.7                             & 85.3                           & 20.0                    & 64.0                    & 37.0                    & 74.7                   & 77.0                    & 95.6                      & 56.0                & 70.4    \\\midrule
\textbf{Avg.}     & \textbf{97.9}                           & {95.4}                             & {88.1}                          & {32.1}                    & {74.6}                    & {52.2}                   & {83.3}                 & {84.9}                   & {97.4}                      & {72.5}               & {82.6}      \\
\bottomrule
\end{tabular}
\end{table*}

From the comparative results presented in Table~\ref{tab:FullComp}, it appears that the results are consistent between the three machine learning tasks. Consequently, the impact of the different parameters and assumptions can be discussed at the general level.

\textbf{DeSpaWN outperforms the baseline.} On all tasks the proposed architecture, DeSpaWN significantly outperforms the baseline models found in the literature. Particularly on MIMII dataset, it reaches globally a performance improvement of 16\%. Also, compared to the baseline, DeSpaWN is much less impacted by the noise level, when the SNR changes from 6 to 0 dB, DeSpaWN experiences a drop of performance of 4\% while the baseline has a 9\% drop. When the noise level increases to a SNR at -6dB, DeSpaWN performances diminishes while remaining well above the baseline (+16\%). This suggests that DeSpaWN is learning a noise-independent representation of the signals, which makes it much more robust to noise than other approaches. This is a particularly important requirement in real applications that are typically impacted by different types of noise at different levels. 
Last, in the baseline, the log-mel spectrogram is extracted to be used as input to an auto-encoder. With DeSapWN, the raw data are used directly, without requiring any pre-processing step. This lightens the methodology significantly since extracting a spectrogram requires the choice of several hyper-parameters such as: the window type, the window length, the window stride, whether to compute the density or the magnitude and whether to apply additional transformations (log-mel, decibels, etc.). All these choices can influence the results significantly. 

\textbf{Impact of the Sparsity Coefficient.} 
From the first three columns of  Table~\ref{tab:FullComp}, it appears that the sparsity term in the loss of DeSpaWN (\eqref{eqn:loss}) influences the results. In addition, setting $\gamma$ to one, without fine-tuning, always seems to be a near optimal choice. This can be explained by our definition of the loss as the average of the $\ell_1$ values, first of the residuals, and second of the coefficients modulus. Using the average makes them comparable. Using smaller $\gamma$ influences the results slightly. However, increasing $\gamma$ can have quite a strong impact on the performance. This signifies that, even though sparsity is helping to get some robustness to external factors, too much of it would be at the expense of the reconstruction loss and at the expense of the ability to distinguish variations in the signals, including anomalies or class specific coefficient behaviours.

\textbf{Impact of Hard Threshold Learning.} 
The second noteworthy observation is that the architectures without hard-threshold are performing significantly worse compared to others (\textit{db4} versus \textit{db4}+HT, or LCWN versus DeSpaWN). This highlights the importance of the denoising part of the architectures. The strength of the architecture is its ability to learn the best thresholds for the wavelet coefficients in order to become robust to small variations in the signal. This strength is independent of whether the wavelets are learnt or not.

\textbf{Impact of Wavelet Learning.} 
For the classification task in particular, learning the right wavelet is pivotal for the architecture accuracy. This is to be expected since the class attribution is done based on the network loss minimisation. When fixing the wavelets or even both the wavelets and the thresholding function (\textit{db4} and \textit{db4}+HT), there are not many parameters left to optimise. The architectures become generic and not fine-tuned per class, making the class attribution based on the loss not much better than a random guess (\textit{db4} has random guess attribution since all architectures are identical). For anomaly detection, this effect is mitigated by the property of the wavelets: thanks to the CQF property, wavelets are designed for good reconstruction and relative sparsity (wavelets are used in signal processing since they inherently tend to produce sparse signal representations). Hence, they are already good candidates to create relevant signal description and, thus, anomaly detection. Yet, learning the right wavelets still brings some non-negligible improvements (additional +1 or 2 \%, averaged over several tens of experiments).

\textbf{Impact of the CQF.} 
The impact of constraining the wavelet basis can be observed by comparing \textit{db4}+HT, (fixed Daubechies-4 wavelets), the DeCWN, (learning of one global wavelet), the DeSpaWN, (learning one wavelet per layer), the DeSpaWN-2, (learning one wavelet and one scaling function per layer) and the FreeWN (learning all kernels). As expected, constraining the kernels tends to make the network less specific to the characteristics of the training class and affects the classification performance strongly. The most constrained architecture (\textit{db4}+HT) has the worst results (52\%), DeCWN is performing better (84.9\%) but not as good as DeSpaWN or DeSpaWN-2 (97.9 and 97.4\%). These two architectures are in fact quite equivalent in terms of results. Leaving the kernel completely free (FreeWN) also leads to a drop in performance likely due to training instabilities as explained in Section~\ref{sec:methods}.

    \subsection{Additional Diagnostics Potential}
    %------------------------------------------------------
        \subsubsection{Insights on MIMII}
        %```````````````````````````````````````````````````````````
        
\begin{figure}
    \centering
    \includegraphics[width=\imwidth]{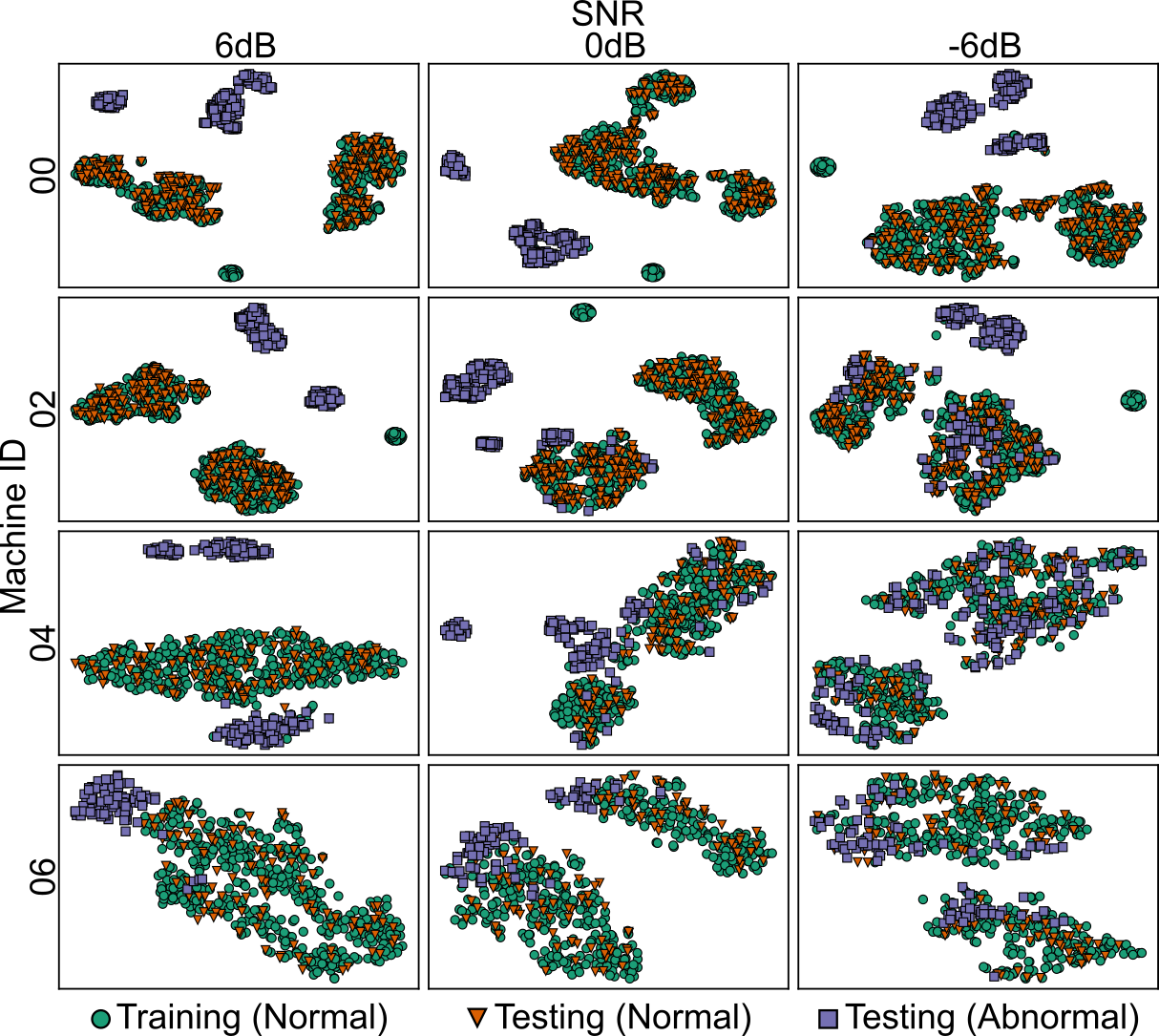}
    \caption{\textbf{T-SNE of the Slide Rail latent space}. Representation in two dimensions of the Slide Rail latent space for the four machines (one machine per row), at the three different experimental SNR (columns), using T-SNE (perplexity of 30). One can distinguish different clusters, most likely representing different operating conditions and anomalies.
    \label{fig:mimiiTsne}}
\end{figure}

\begin{figure*}
    \centering
    \includegraphics[width=12cm]{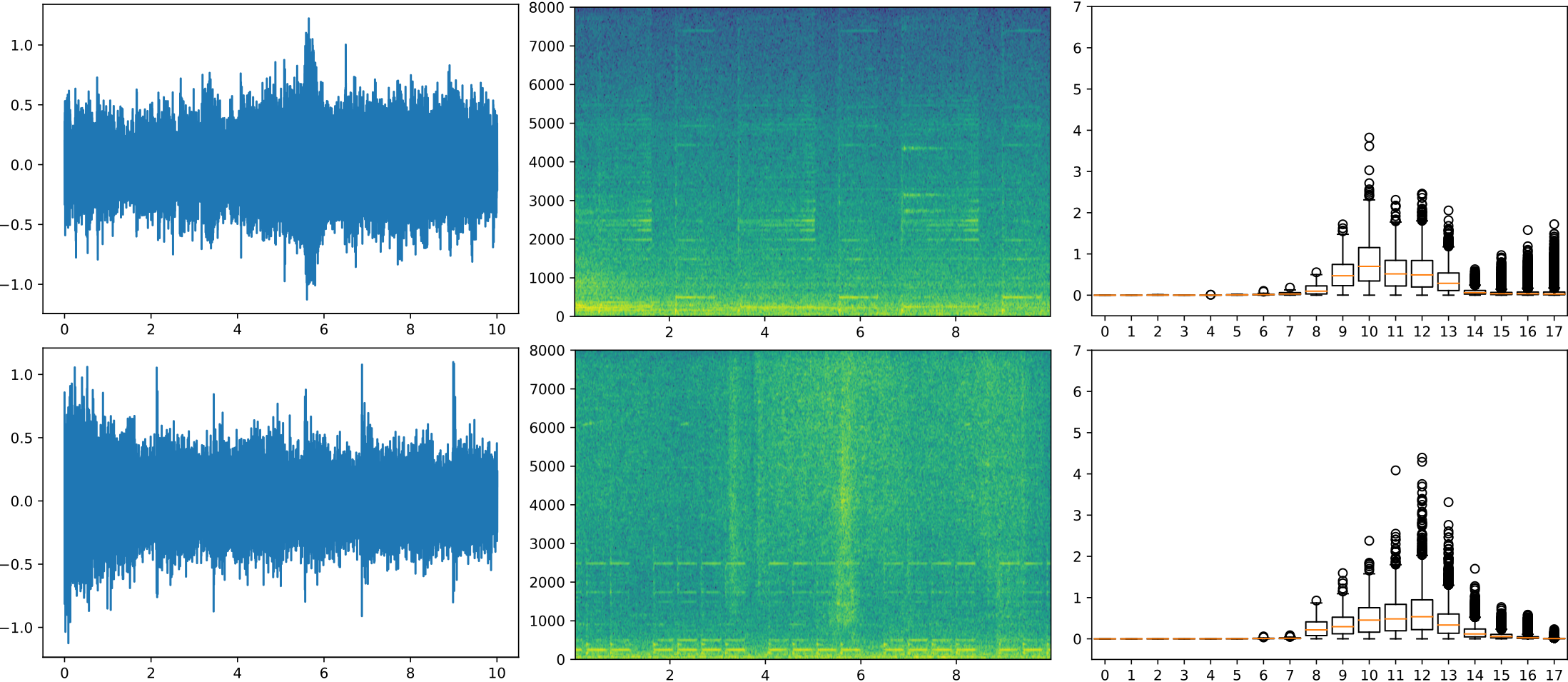}
    \caption{\textbf{Examples of Normal Signals}. Raw data, log-spectrogram and obtained coefficient distribution per level for two normal measurements of the Slider Rail 0 at SNR 0dB.
    \label{fig:healthExample}}
\end{figure*}

\begin{figure*}
    \centering
    \includegraphics[width=12cm]{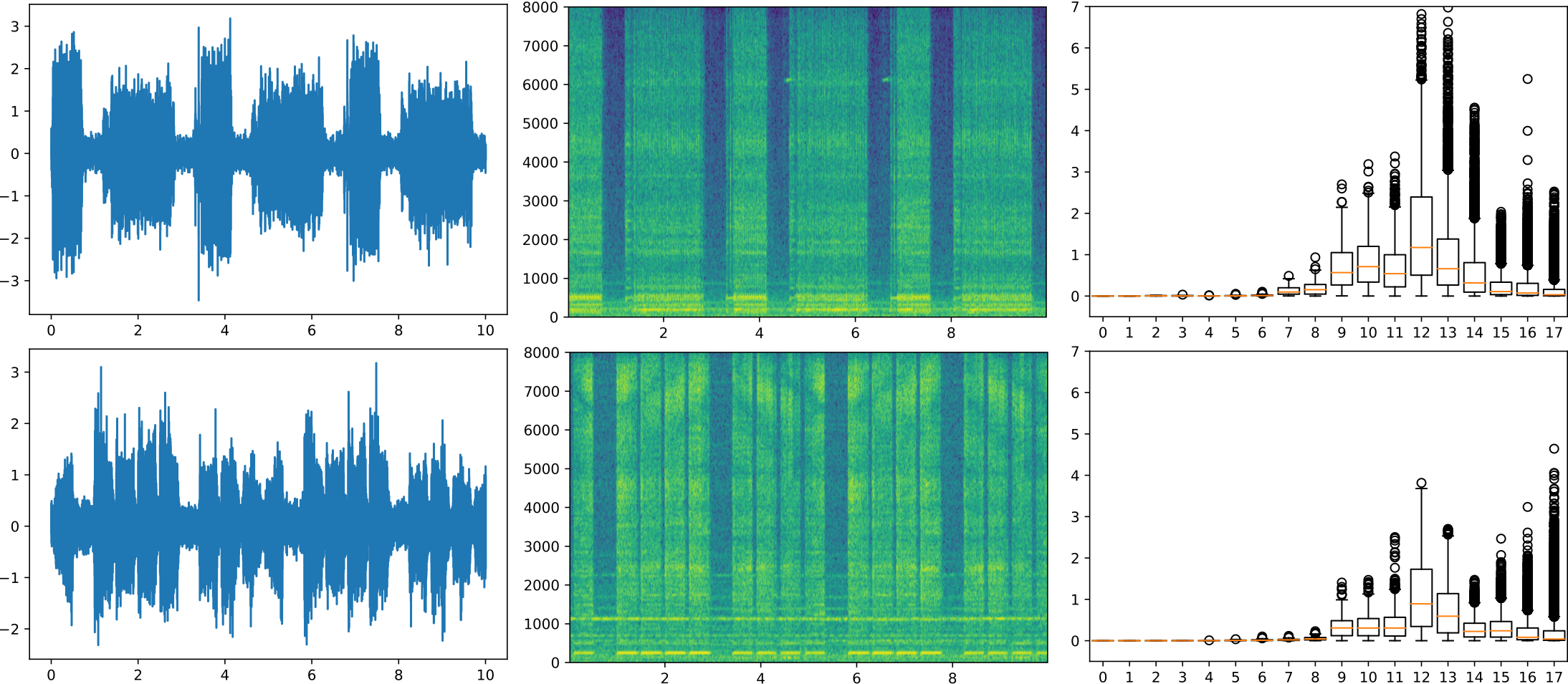}
    \caption{\textbf{Examples of Abnormal Signals}. Raw data, log-spectrogram and obtained coefficient distribution per level for two abnormal measurements of the Slider Rail 0 at SNR 0dB.
    \label{fig:unhealthExample}}
\end{figure*}

\begin{figure}
    \centering
    \includegraphics[width=\imwidth]{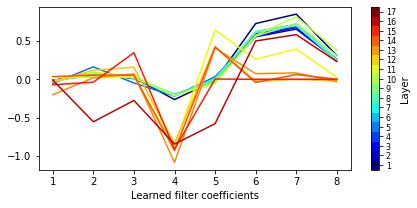}
    \includegraphics[width=\imwidth]{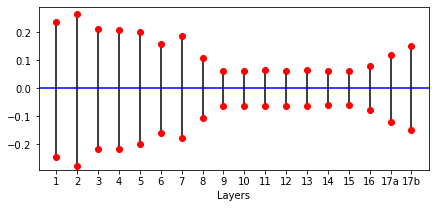}
    \caption{\textbf{Learned coefficients and biases}. (top) Learned kernel for all layer of a Slider Rail. Each color represent the filter from a different layer, from first (high-frequency) layer in dark blue to the last layers in red. (bottom) Learned positive and negative biases for each layer. The black lines represent the range of values that are set to zeros by the corresponding  hard-thresholding layer.
    \label{fig:mimiiKernel}}
\end{figure}

The good detection performance of DeSpaWN indicates again that the signals are probably forming well defined clusters in the $Res-MaxRes-\lbrace\hat{\ell}^l_1\rbrace-\lbrace\bar{\ell}^l_1\rbrace_{l\in[1..L]}$ latent space. Anomalies would then appear outside of these clusters and finding anomaly clusters could help to diagnose the different conditions of the system. This can be visualised in two dimensions, by performing a t-distributed stochastic neighbor embedding (t-SNE) on the latent space, as illustrated in Figure~\ref{fig:mimiiTsne}. In this figure, the latent spaces of the different slide rail experiments are shown after a t-SNE transformation with default perplexity of 30. Clusters can be clearly identified in this representation. They are likely to be formed by different anomaly types and operating conditions. In all experiments, one can distinguish at least two normal operating condition clusters, indicating different conditions of operation and at least two anomaly clusters, well separated from the normal conditions. With this representation, only when the SNR decreases to -6dB, some of the anomalous points become less separated from the normal conditions. This decrease in separability could be expected from the lower AUC as reported in Table~\ref{tab:FullComp}. 

The diagnostics possibilities offered by extracting characteristic patterns of the learned coefficients of the proposed approach for the different clusters analysis  are illustrated for the Slide Rail 0 at 0dB SNR in Figure~\ref{fig:healthExample}. Two different signals extracted from each of the two normal measurement clusters are shown. In this Figure, the raw signals, their log-spectrogram and the distribution of the learned coefficients over the 18 levels are shown. The first signal has its major components around the 10th level with quite some large coefficients at the highest level of detail, while the other signal is mostly ``active'' at the 12th level, with very little activation of the last two levels (highest level of details). This indicates that the operating mode has likely changed between these two samples: the main information content changed from level 10 to 12, that is, a factor 4 in the spectrum of the original signal. Similarly, on Figure~\ref{fig:unhealthExample}, two unhealthy signals of the slider rail drawn  from two different clusters are depicted. The first signal has its 12th level more activated than healthy signals; the other signal distinguishes itself with its much larger high level coefficients. These are likely two different types of anomalies.

Last, Figure~\ref{fig:mimiiKernel} shows exemplarily the learned filters and hard-threshold coefficients for a Slider Rail. The first layers, corresponding to the high-frequencies, have high hard-thresholding values (up to $0.3$ in absolute value). One can observe that the corresponding wavelet coefficients in Figure~\ref{fig:healthExample} are almost all zero. This makes the filter of these layer irrelevant and this further explains why the corresponding filters observed in Figure~\ref{fig:mimiiKernel} are very close to the original Daubechie filters. It is probably where most of the noise is concentrated. For lower frequencies, the filters are further away from the Daubechie wavelets and the hard-thresholding is much lower. It is probably at these scales that the information required for the reconstruction is concentrated. This interpretation matches the coefficient distributions observed in Figure~\ref{fig:healthExample}. This gives a strong indication that the proposed architecture can indeed selectively filter and threshold the layers based on their information content.

    \subsubsection{Insights on the Bird Song Dataset}
    %----------------------------------------------
\begin{figure*}
    \centering
    \includegraphics[width=\textwidth]{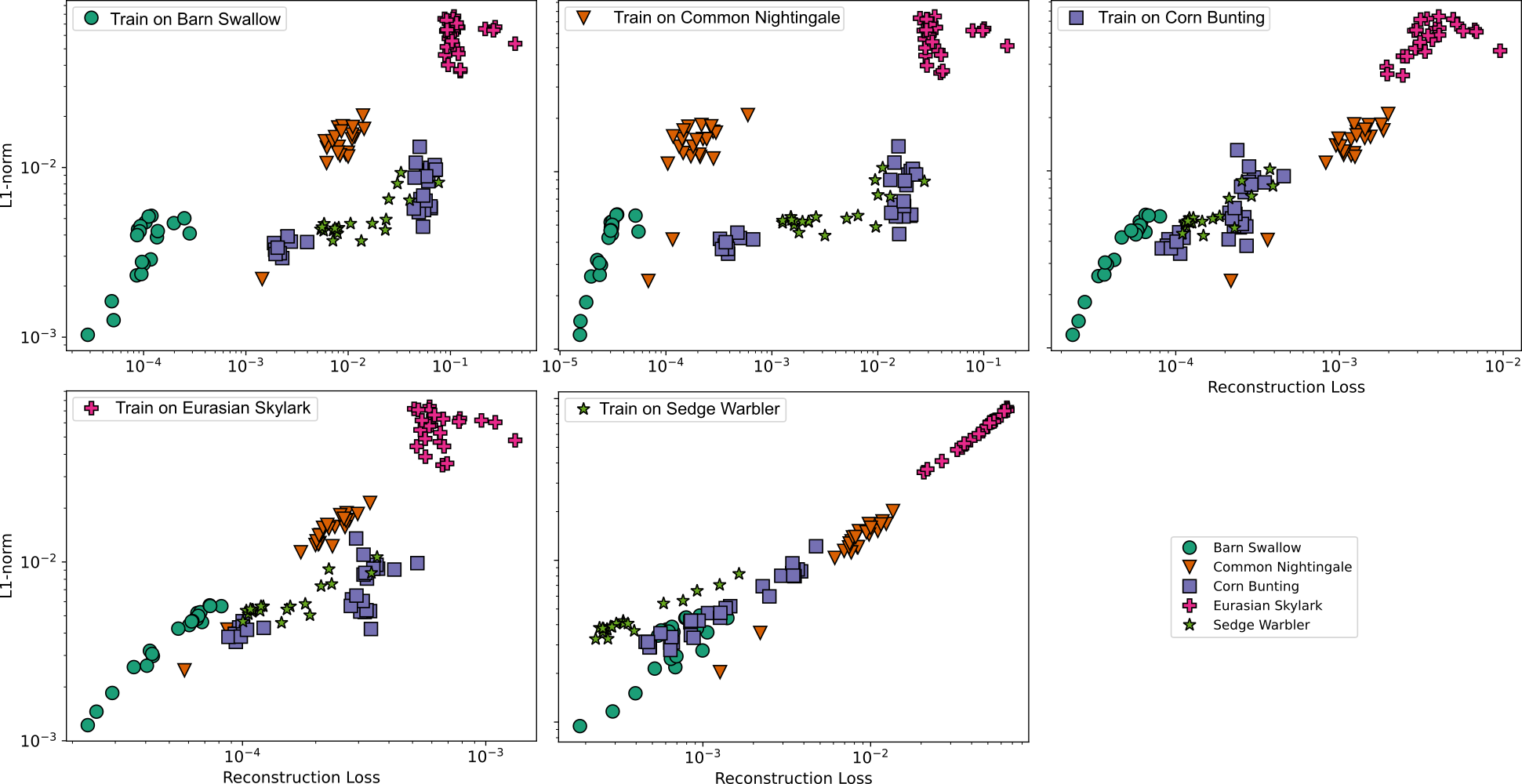}
    \caption{\textbf{$Res-\hat{\ell}_1$ Space}. For each bird, DeSpaWN is trained and all birds are tested and plotted together. When DeSpaWN is trained on one species, the species tends to be well separated from others, and has generally a more compact representation in the $Res-\hat{\ell}_1$ than when tested with another DeSpaWN.
    \label{fig:BirdAnom}}
\end{figure*}

Figure~\ref{fig:BirdAnom} illustrates the $l1$-residual space for all birds for the different cases. For each plot, DeSpaWN is trained on the main class bird where all other bird samples would have to be detected as anomalies. In each case, one can observe that the samples corresponding to the bird used for training form a well-defined and separated cluster, allowing the one-class classifier to identify easily all other birds as anomalies. The only exception is the Corn Bunting case (third plot), which is mixed with the Sedge Warbler after training, particularly with this specific representation. These results explain the classification accuracy drop observed when adding the Sedge Warbler as the fifth-species (Table~\ref{tab:FullComp}).

%%%%%%%%%%%%%%%%%%%%%%%%%%%%%%%%%%%%%%%%%%%%%%%%%%%%%%%%%%%%%%%%%%%%%%%%%
\section{Comparison to other frameworks}
\label{sec:compOther}
%%%%%%%%%%%%%%%%%%%%%%%%%%%%%%%%%%%%%%%%%%%%%%%%%%%%%%%%%%%%%%%%%%%%%%%%%

\begin{table*}
\centering
\caption{\textbf{Comparative Study on three machine learning tasks:} For the different methodologies (one per column), comparative results on the three considered tasks. For the anomaly detection tasks, on MIMII and on the Bird Song, we report the average AUC (\%).%
\label{tab:ComptoLit}}
\begin{tabular}{@{}crrrrrrrrrrr@{}}
\toprule
         & \multicolumn{1}{c}{DeSpaWN}& 
         \multicolumn{2}{c}{Scattering transform}    & \multicolumn{4}{c}{Auto Encoder} & 
         \multicolumn{4}{c}{Unet} \\ 
         & \multicolumn{1}{c}{$\gamma$} & \multicolumn{1}{c}{$(J,Q)$} & \multicolumn{1}{c}{$(J,Q)$} & \multicolumn{1}{c}{$N_{AE}$} & \multicolumn{1}{c}{$N_{AE}$} & \multicolumn{1}{c}{$N_{AE}$} & \multicolumn{1}{c}{$N_{AE}$} & \multicolumn{1}{c}{$N_{Unet}$}  & \multicolumn{1}{c}{$N_{Unet}$}  & \multicolumn{1}{c}{$N_{Unet}$} &  \multicolumn{1}{c}{$N_{Unet}$}\\ 
         & \multicolumn{1}{c}{$1$} & \multicolumn{1}{c}{$(17,1)$} & \multicolumn{1}{c}{$(10,8)$} & \multicolumn{1}{c}{$4$} & \multicolumn{1}{c}{$8$} & \multicolumn{1}{c}{$16$} & \multicolumn{1}{c}{$32$} & \multicolumn{1}{c}{$1$}  & \multicolumn{1}{c}{$2$}  & \multicolumn{1}{c}{$4$} &  \multicolumn{1}{c}{$8$}\\ \midrule
         & \multicolumn{11}{c}{Anomaly Detection on MIMII}                                                                                                                                                                                                                                                      \\\midrule
Valve    & \textbf{92.8}                           & 65.5                             & 78.6                           & 91.5                    & 92.4                    & 91.5                    & 92.0                   & 92.3                    & 92.7                     & 93.0                & 91.1    \\
Pump     & 84.5                           & 88.5                            & \textbf{90.6}                           & 74.5                    & 74.5                    & 74.0                    & 73.4                   & 75.4                    & 69.6                      & 70.5                & 74.4    \\
Fan      & 86.2                          & \textbf{88.7}                            &84.9                       & 71.0                   & 76.6                    & 80.7                    & 76.1                   & 77.1                    & 75.6                      & 76.5                &  73.6   \\
Slider   & 91.0                           & 86.9                             & \textbf{96.8}                           & 79.2                    & 81.0                    & 81.7                    & 78.1                   & 78.6                    & 80.6                      & 83                & 87.4    \\
6 dB     & 94.6                           & 91.2                           & \textbf{95.4}                           & 83.4                    & 87.5                    & 86.6                    & 85.0                   & 86.3                   & 84.0                      & 84.7               & 85.2    \\
0    dB  & \textbf{90.5}                           & 84                            & 88.6                           & 79.7                    & 81.3                    & 81.5                    & 80.8                   & 82.5                    & 79.8                      & 83.0                & 83.8    \\
-6    dB & \textbf{80.8}                           & 72                             & 79.1                        & 74.2                   & 74.6                    & 73.9                    & 74.0                   & 74.0                    & 75.0                      & 74.0                 & 75.0   \\ \midrule
\textbf{Avg.}     & \textbf{88.6}                           & {82.4}                             & {87.7}                           & {79.1}                    & {81.15}                   & {80.3}                    & {80.0}                   & {81.0}                    & {79.6}                      & \textbf{80.9}              & {81.4}      \\ \midrule\midrule
         & \multicolumn{11}{c}{Anomaly Detection (1 versus 4 bird species)}                                                                                                                                                                                                                           \\\midrule
{Avg.}      & \textbf{98.6}                           & {93.8}                             & {94.0}                           & {85.4}                    & {86.1}                    & {86.3}                   & {86.3}                   & {89.6}                    & {88.9}                      & {89.2}              &  {90.0}      \\\midrule\midrule
\#C      & \multicolumn{11}{c}{Classification with SVM (BirdSong)}                                                                                                                                                                                                                                             \\\midrule
{5}         & 97.7                           & \textbf{97.9}                             & 97.3                           & 87.6                    & 92.2                    & 92.3                    & 90.2                   & 83.4                    & 85.7                      & 87.1                & 88.4   \\
\bottomrule
\end{tabular}
\end{table*}

\subsection{Scattering Transform, U-Net and Auto-Encoders}

In this section, we propose to compare the results of DeSpaWN to other sate-of-the art frameworks found in the literature: the scattering transform, vanilla convolution auto-encoders (CAE) and U-net.

First, we propose to compare to the scattering transform \cite{anden2014deep}, which has been extensively used in the context of audio signal processing. It is a signal representation that is stable to small deviations in its inputs, and which is able to characterize transient phenomena like amplitude modulation. We use the Kymatio library \cite{andreux2020kymatio} to compute the coefficients from a two-layer scattering transform. It requires the selection of two parameters \cite{destouet2020wavelet}: $J$ the maximum scale of the filters used, implying that the transform will only capture frequencies superior to $2^J$, and $Q$ the number of filters per octave. We propose to analyse two combinations of parameters: 1) $J=17$ and $Q=1$ which will result, for the first layer, in a decomposition close to the FDWT with one filter per octave and able to characterize low frequencies up to $2^{-17}$; and 2)  $J=10$ and $Q=8$ for the first layer (which defines wavelets having the same frequency resolution as Mel-frequency filters), and $Q=1$ for the second layer. The second choice of parameters is motivated by a previous research that proposed to consider Mel-spectrogram features on frames of around 60 milliseconds~\cite{purohit2019mimii}.
Similarly to the approach used with DeSpaWN, for the anomaly detection task, we use an One-Class HELM on the scattering coefficients. For the Bird Song classification task, since the scattering transform does not create signal specific decomposition, the dictionary learning approach cannot be mimicked. Therefore, we use, for all approach,  the coefficient as input to an SVM classifier to make the results comparable.

In addition, to highlight the relevance of the proposed architectural choice of DesPaWN, we compare it numerically to standard CNN autoencoders (CAE) and CNN U-nets. The considered AE autoencoders are based on the work~\cite{liu2019fault}. The architecture has been used for fault diagnosis of rotating machinery. We consider 4 encoding and decoding layers with 8 coefficients per kernel and a kernel size of $N_{AE}$ for each layer. The impact of the addition of trainable parameters is studied by considering the range of $N_{AE}=[4,8,16,32]$. 
DeSpaWN architecture has skip-connections at each level. Therefore, it can be considered as a special case of a U-net model. We then compare our method to another U-net architecture based on \cite{jimenez2019u} which was applied to electrocardiogram detection. We replace the concatenation of the skipped connection with the addition in order to have an architecture closer to DesPaWN. Furthermore, we consider $L=17$ layers with a stride of 2 for each CNN layer and 8 coefficients per kernel. The kernel size $N_{Unet}$ of each filter is studied in the range $N_{Unet}=[1,2,4,8]$. This grid was chosen since larger kernel sizes led to decreasing performances. The main differences between our model and this U-net architecture are: one more filter at each skip connection, initialisation of each pair of filters as high-pass and low-pass, and replacing ReLu activation function with the proposed learnable Hard-Thresholding function.
For each method, the same loss function as in \eqref{eqn:loss} is used for training. The exact same process is followed for anomaly detection and for classification as for the results achieved with DeSpaWN (\cf Section~\ref{sec:classifAnomAppro}).

\subsection{Results of the Benchmark}
    %--------------------------------

The results of the benchmark are presented in Table~\ref{tab:ComptoLit}, for two versions of the scattering transform, four variations of the CAE and four variations of the U-net.
DeSpaWN and the scattering transform both provide very competitive results and are both very solid candidates to solve the tasks on the dataset studied here. One can note, however, a bigger drop in performance for the scattering transform when the noise on the data increases (-20\% and -16\% when comparing MIMII 6dB with -6dB). It is also worth noting that the strength of each approach depends on the machine type, where the scattering transform seems the most adapted to tackle the pump and the slider while DeSpaWN is better performing on the valve and on the fan. For the Bird Song classification, very similar results are achieved.

The auto-encoders, both the traditional CAE and the U-nets, provide very competitive results for the valve system but not for the other machines and are overall significantly less performing compared to the scattering transform and to DeSpaWN. It is worth noting, however, that all approaches outperform the reported baseline on these datasets.

%%%%%%%%%%%%%%%%%%%%%%%%%%%%%%%%%%%%%%%%%%%%%%%%%%%%%%%%%%%%%%%%%%%%%%%%%
\section{Conclusion}
%%%%%%%%%%%%%%%%%%%%%%%%%%%%%%%%%%%%%%%%%%%%%%%%%%%%%%%%%%%%%%%%%%%%%%%%%

In this paper, we proposed an architecture for learning a meaningful and sparse representation of high-frequency signals in an unsupervised manner without requiring neither pre-processing (feature extraction) nor post-processing (\eg denoising). This architecture achieves very good results that are  well above the baselines and are competitive compared to other state-of-the-art approaches on three machine-learning tasks for anomaly detection and classification. We designed an end-to-end deep learning architecture, mimicking the cascade algorithm of the FDWT but making it fully learnable. Using the deep-learning framework, we demonstrated the benefits of learning the right wavelets at each level of the decomposition. One of the additional contributions is the introduction of a learnable hard-thresholding function for automatic signal denoising.

The proposed methodology combines a thorough theoretical foundation on the wavelet properties, including cascade, perfect reconstruction and anti-aliasing filter basis with the CQF property, denoising with coefficient thresholding with the learning ability of deep learning. The proposed architecture could demonstrate a significant improvement on sound data processing, both for classification and for anomaly detection tasks. Our approach allows the use of the raw HF data as input to a deep learning architecture, a setup usually avoided in the literature due to the difficulty of designing efficient architectures that are robust to changes in the input lengths. The proposed architecture takes root in spectral analysis and can replace the usual pre-processing steps such as spectrogram or wavelet coefficient extraction. Since it is unsupervised, it can be used as an input to subsequent learning methods. In addition, compared to other deep learning architectures, it is a very light framework with only few hundreds of learnable parameters, mitigating in that way the high risk of overfitting. With its spectral interpretation, it also provides diagnostics information to the domain experts that can potentially improve the interpretation capabilities.

This work opens several doors for future directions. First, given the high information content of the proposed latent space, other unsupervised machine learning tasks could be explored such as system degradation monitoring. \Eg, a drop in the sparsity of the decomposition could be a sign of an increased signal complexity or of the presence of disturbing components due to system wear. The architecture could be further analysed in conditions with controlled noise and signals to better understand its denoising and stability properties. The architecture could also be extended, such as in particular with the imbrication of this architecture in stacked architecture to solve supervised machine learning tasks. In these cases, the learned wavelets and thresholding coefficients could be learned not only for sparsity and for reconstruction but on top of that to maximise a supervised objective. The use of parallel wavelets (number of kernels in a convolution layer), or the handling of multi-channel inputs are further exciting potential developments. 

    \section*{Acknowledgment}
        \aknow
    \bibliographystyle{abbrvnat}
    \bibliography{references}

\end{document}